\documentclass[11pt]{article}
\usepackage{eamt24}
\usepackage{times}
\usepackage{url}
\usepackage{latexsym}
\usepackage[small,bf]{caption} 
\usepackage{booktabs}
\usepackage{multicol}
\usepackage{multirow}
\usepackage{verbatim}
\usepackage{graphicx}
\usepackage{subfigure}
\usepackage{amsmath}
\usepackage{amssymb}

\graphicspath{ {./} }

\title{A Case Study on Context-Aware Neural Machine Translation with Multi-Task Learning}

\author{Ramakrishna Appicharla$^1$, Baban Gain$^1$, Santanu Pal$^2$, Asif Ekbal$^3$, Pushpak Bhattacharyya$^4$ \\ $^1$Department of Computer Science and Engineering, Indian Institute of Technology Patna, India \\ $^2$Wipro AI, Lab45, London, UK \\ $^3$School of AI and Data Science, Indian Institute of Technology Jodhpur, India \\ $^4$Department of Computer Science and Engineering, Indian Institute of Technology Bombay, India \\ {\tt \{ramakrishnaappicharla, gainbaban, santanu.pal.ju,} \\ {\tt asif.ekbal, pushpakbh\} @gmail.com}}

\date{}

\begin{document}
\maketitle
\begin{abstract}
  In document-level neural machine translation (DocNMT), multi-encoder approaches are common in encoding context and source sentences. Recent studies \cite{li-etal-2020-multi-encoder} have shown that the context encoder generates noise and makes the model robust to the choice of context. This paper further investigates this observation by explicitly modelling context encoding through multi-task learning (MTL) to make the model sensitive to the choice of context. We conduct experiments on cascade MTL architecture, which consists of one encoder and two decoders. Generation of the source from the context is considered an auxiliary task, and generation of the target from the source is the main task. We experimented with German--English language pairs on News, TED, and Europarl corpora. Evaluation results show that the proposed MTL approach performs better than concatenation-based and multi-encoder DocNMT models in low-resource settings and is sensitive to the choice of context. However, we observe that the MTL models are failing to generate the source from the context. These observations align with the previous studies, and this might suggest that the available document-level parallel corpora are not context-aware, and a robust sentence-level model can outperform the context-aware models.
\end{abstract}

\section{Introduction}
\label{sec: introduction}

Context-aware neural machine translation gained much attention due to the ability to incorporate context, which helps in producing more consistent translations than sentence-level models \cite{maruf-haffari-2018-document,zhang-etal-2018-improving,bawden-etal-2018-evaluating,agrawal2018contextual,voita-etal-2019-good,huo-etal-2020-diving,li-etal-2020-multi-encoder,donato-etal-2021-diverse}. There are mainly two approaches to incorporating context. The first one is to create a context-aware input sentence by concatenating context and current input sentence \cite{tiedemann-scherrer-2017-neural,agrawal2018contextual,junczys-dowmunt-2019-microsoft,zhang-etal-2020-long} and using it as the input to the encoder. The second approach uses an additional context-aware component to encode the source or target context \cite{zhang-etal-2018-improving,voita-etal-2018-context,kim-etal-2019-document,ma-etal-2020-simple} and the entire model is jointly optimized. Typically, the current sentence's neighbouring sentences (either previous or next) are used as the context.

The context-aware models are trained to maximize the log-likelihood of the target sentence given the source sentence and context. Most of the existing works on DocNMT \cite{zhang-etal-2018-improving,maruf-haffari-2018-document,voita-etal-2019-good,li-etal-2020-multi-encoder} focus on encoding the context through context-specific encoders. Recent studies \cite{li-etal-2020-multi-encoder} show that, in the multi-encoder DocNMT models, the performance improvement is not due to specific context encoding but rather the context-encoder acts like a noise generator, which, in turn, improves the robustness of the model. In this work, we explore whether the context encoding can be modelled explicitly through multi-task learning (MTL) \cite{luong2015multi}. Specifically, we aim to study the effectiveness of the MTL framework for DocNMT rather than proposing a state-of-the-art system. The availability of document-level corpora is less compared to sentence-level corpora. Previous works \cite{junczys-dowmunt-2019-microsoft} use the sentence-level corpora to warm-start the document-level model, which can be further tuned with the existing limited amount of document-level data. However, in this work, we focus only on improving the performance of DocNMT models with available document-level corpora. We consider the source reconstruction from the context as the auxiliary task and the target translation from the source as the main task. We conduct experiments on cascade MTL \cite{anastasopoulos-chiang-2018-tied,zhou-etal-2019-improving} architecture. The cascade MTL architecture comprises one encoder and two decoders (Figure \ref{fig:model}). The intermediate (first) decoder attends over the output of the encoder, and the final (second) decoder attends over the output of the intermediate decoder. The input consists of $\langle\mathrm{c_x}, \mathrm{x}, \mathrm{y}\rangle$ triplets, where $\mathrm{c_x}$, $\mathrm{x}$ and $\mathrm{y}$ represents the context, source, and target sentences, respectively. The model is trained to optimize both translation and reconstruction objectives jointly. We also train two baseline models as contrastive models, namely sentence-level vanilla baseline and single encoder-decoder model, by concatenating the context and source \cite{tiedemann-scherrer-2017-neural,agrawal2018contextual,junczys-dowmunt-2019-microsoft}. We additionally train multi-encoder single-decoder models \cite{li-etal-2020-multi-encoder} to study how context affects the DocNMT models. We conduct experiments on German-English direction with three different types of contexts (\textit{viz.} previous two source sentences, previous two target sentences, and previous-next source sentences) on News-commentary v14 and TED corpora. We report BLEU \cite{papineni-etal-2002-bleu} calculated with sacreBLEU \cite{post-2018-call} and APT (accuracy of pronoun translation) \cite{miculicich-werlen-popescu-belis-2017-validation} scores.

To summarize, the specific attributes of our current work are as follows:

\begin{itemize}
    \item We explore whether the MTL approach can improve the performance of context-aware NMT by introducing additional training objectives along with the main translation objective.
    \item We propose an MTL approach where the reconstruction of the source sentence given the context is used as an auxiliary task and the translation of the target sentence from the source sentence as the main task, jointly optimized during the training.
    \item The results show that in the MTL approach, the context encoder generates noise similar to the multi-encoder approach \cite{li-etal-2020-multi-encoder}, which makes the model robust to the choice of the context.
\end{itemize}

\section{Related Work}
\label{sec: related_works}

Previous studies have proposed various document-level NMT models and achieved great success. The main goal of these approaches is to efficiently model context representation, which can lead to better translation quality.
Towards this goal to represent context,  Tiedemann and Scherrer~\shortcite{tiedemann-scherrer-2017-neural} concatenate consecutive sentences and use them as input to the single-encoder-based DocNMT model. Agrawal et al.~\shortcite{agrawal2018contextual} conducted experiments on varying neighbouring contexts and concatenated with the current sentence as input to their model. With these similar trends, Junczys-Dowmunt~\shortcite{junczys-dowmunt-2019-microsoft} conducted experiments considering the entire document as context. Further progress on context representation in DocNMT, Zhang et al.~\shortcite{zhang-etal-2018-improving} and Voita et al.~\shortcite{voita-etal-2018-context} proposed transformer-based multi-encoder NMT models where the additional encoder is used to encode the context. While Miculicich et al.~\shortcite{miculicich-etal-2018-document} proposed a hierarchical attention network to encode the context, a more recent approach Kang et al.~\shortcite{kang-etal-2020-dynamic} proposed a reinforcement learning-based dynamic context selection module for DocNMT. Kim et al.~\shortcite{kim-etal-2019-document} and Li et al.~\shortcite{li-etal-2020-multi-encoder} conducted experiments on multi-encoder DocNMT models and reported that the performance improvement is not due to context encoding; instead, the context encoder acts as a noise generator, which improves the robustness of the DocNMT model. Junczys-Dowmunt~\shortcite{junczys-dowmunt-2019-microsoft} conducted experiments on a single encoder model with masked language model objective \cite{devlin-etal-2019-bert} to incorporate document-level monolingual source-side data. Since the multi-encoder models are trained to optimize the translation objective only, it might be possible for the model to pay less attention to the context, and Li et al.~\shortcite{li-etal-2020-multi-encoder} report the same.

MTL strategies in NMT trained on other auxiliary tasks along with the main translation task \cite{luong2015multi,dong-etal-2015-multi,zaremoodi-etal-2018-adaptive,wang-etal-2020-multi,yang2020towards} achieved significant improvements in translation quality so far. The other auxiliary tasks include autoencoding \cite{luong2015multi}, denoising autoencoding \cite{wang-etal-2020-multi}, parsing and named entity recognition \cite{zaremoodi-haffari-2018-neural,zaremoodi-etal-2018-adaptive}. Zhou et al.~\shortcite{zhou-etal-2019-improving} proposed a cascade MTL network to improve the robustness of the NMT model. They considered denoising the noisy text as an auxiliary task and the translation as the main task. They achieved a significant BLEU score improvement (up to 7.1 BLEU) on the WMT robustness shared task on the French-English dataset.

However, most multi-task models are proposed only for sentence-level NMT models. Multi-task learning is relatively unexplored in context-aware NMT settings. Wang et al.~\shortcite{wang-etal-2021-autocorrect} proposed an MTL framework for dialogue translation tasks that jointly correct the sentences having issues such as pronoun dropping, punctuation dropping, and typos and translate them into the target language. Liang et al.~\shortcite{liang-etal-2022-scheduled} proposed a three-stage training framework for the neural chat translation task. The model is trained on auxiliary tasks such as monolingual cross-lingual response generation tasks to generate coherent translation and the next utterance discrimination task. Lei et al.~\shortcite{lei-etal-2022-codonmt} proposed an MTL system to force the model to attend over relevant cohesion devices while translating the current sentence. In this work, we propose a multi-task learning objective, i.e., reconstruction of source sentences given the source context in a cascade multi-task learning setting to study the effect of context in document-level NMT systems.

\section{Methodology}
\label{sec: methodology}

\subsection{Problem Statement}
Our document-level NMT is based on a cascade MTL framework to force the model to consider the context while generating translation. Given a source sentence $\mathrm{x}$ and context $\mathrm{c_x}$, the translation probability of the target sentence $\mathrm{y}$ in the DocNMT setting is calculated as in Equation~\ref{eq:trans_obj}.

\begin{equation}
    \label{eq:trans_obj}
    p(\mathrm{y}) = p(\mathrm{y}|\mathrm{x}, \mathrm{c_x}) \times p(\mathrm{x}, \mathrm{c_x})
\end{equation}

We consider $p(\mathrm{x}, \mathrm{c_x})$ as the auxiliary task of source ($\mathrm{x}$) reconstruction from $\mathrm{c_x}$ (as $p(\mathrm{x}|\mathrm{c_x})$) \footnote{Since the joint probability of $p(\mathrm{x}, \mathrm{c_x})$ can be calculated as $p(\mathrm{c_x}|\mathrm{x}) \times p(\mathrm{x})$, we also explored this setting. We observed that the performance of the model is poor in this setting compared to the other setting. More details can be found in Appendix~\ref{sec: aux_objectives}.}, calculated as in Equation~\ref{eq:mtl_obj}.

\begin{equation}
    \label{eq:mtl_obj}
    p(\mathrm{x}, \mathrm{c_x}) = p(\mathrm{x}|\mathrm{c_x}) \times p(\mathrm{c_x})
\end{equation}

The training data $\mathrm{D}$ consists of triplets $\langle\mathrm{c_x}, \mathrm{x}, \mathrm{y}\rangle$. Given the parameters of the model $\theta$, the translation  (Equation~\ref{eq:trans_obj}) and reconstruction (Equation~\ref{eq:mtl_obj}) objectives can be modeled as Equation~\ref{eq:final_obj_trans} and Equation~\ref{eq:final_obj_recons}.

\begin{equation}
    \label{eq:final_obj_trans}
    p(\mathrm{y}|\mathrm{x}, \mathrm{c_x}; \theta) = \prod_{t=1}^{T} p(y_t|\mathrm{x}, \mathrm{c_x}, \mathrm{y}_{<t}; \theta)
\end{equation}

\begin{equation}
    \label{eq:final_obj_recons}
    p(\mathrm{x}|\mathrm{c_x}; \theta) = \prod_{s=1}^{S} p(x_s|\mathrm{c_x}, \mathrm{x}_{<s}; \theta)
\end{equation}
where, $\mathrm{S, Z, T}$ denote the lengths of $\mathrm{x}$, $\mathrm{c_x}$, $\mathrm{y}$ respectively and $\mathrm{x}_{<s}, \mathrm{c_x}_{<z}, y_{<t}$ denote partially generated sequences.

Given translation objective $p(\mathrm{y}|\mathrm{x}, \mathrm{c_x})$ and reconstruction objective $p(\mathrm{x}|\mathrm{c_x})$, the model is jointly trained and optimized the loss, $\mathcal{L}$ using parameter $\theta$ (cf. Equation~\ref{eq5:loss}); where $\alpha$ is a hyper-parameter used to control the loss. We set $\alpha$ to $0.5$.

\begin{equation}
    \label{eq5:loss}
    \begin{split}
    \mathcal{L} = \alpha * \log p(\mathrm{y}|\mathrm{x}, \mathrm{c_x}; \theta) + \\ (1 - \alpha) * \log p(\mathrm{x}|\mathrm{c_x}; \theta)
    \end{split}
\end{equation}

We hypothesize that forcing the model to learn reconstruction and translation objectives jointly will enable the model to encode the context effectively. The output of the reconstruction task can verify this during testing. If the context encoder generates noise, then the model might be unable to reconstruct the source and vice-versa.

\subsection{Cascade Multi-Task Learning Transformer}
\begin{figure}[!ht]
    \centering
    \includegraphics[scale=0.55]{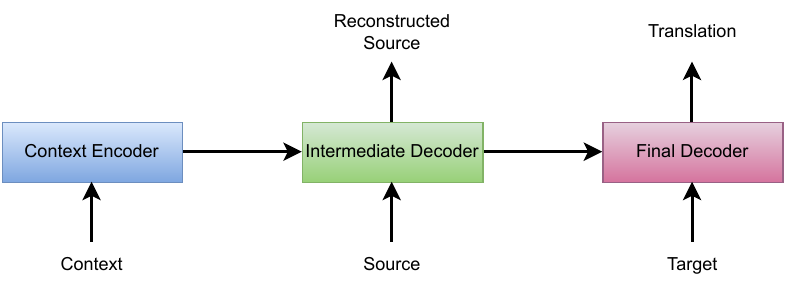}
    \caption{The overview of our MTL architecture. The input to the model is a triplet. The triplet consist of (\textit{Context}, \textit{Source}, \textit{Target}). The Intermediate Decoder is trained to reconstruct the \textit{Source} given \textit{Context}, and the Final Decoder is trained to translate the \textit{Source}. Here, \textit{Source}: Current source sentence, \textit{Context}: Context for the current source sentence, and \textit{Target}: Translation of current source sentence. None of the layers are shared.}
    \label{fig:model}
\end{figure}

The cascade multi-task learning architecture \cite{zhou-etal-2019-improving} (Figure \ref{fig:model}) consists of one encoder and two decoders based on the transformer \cite{vaswani2017attention} architecture. The model takes three inputs: \textit{Source}: Current source sentence, \textit{Context}: Context of the current source sentence, and \textit{Target}: Current target sentence. The input to the encoder is context, and the input to the intermediate decoder is the source. The intermediate decoder is trained to reconstruct the source given context by attending to the output of the encoder. The final decoder attends over the output of the intermediate decoder. In the non-MTL setting, the model is trained only on the translation objective (output of the final decoder), and the intermediate decoder is not trained with the reconstruction objective.

\subsection{Context Selection}
\label{subsec: src_cntx}
We conduct experiments on different settings of the source context. The term ``source context'' is defined as considering related or dependent sentences directly related to the input sentence. Based on the findings of Zhang et al.~\shortcite{zhang-etal-2018-improving}, we select two sentences as context and concatenate them with a special token `$\langle\mathrm{break}\rangle$' \cite{junczys-dowmunt-2019-microsoft}. For a given input source sentence ($\mathrm{x_i}$) and target sentence ($\mathrm{y_i}$), contexts selected for the experiments are:

\begin{itemize}
    \item Previous-2 Source (\textbf{P@2-SRC}): Two previous source sentences ($\mathrm{x_{i-2}, x_{i-1}}$)
    \item Previous-2 Target (\textbf{P@2-TGT}): Two previous target sentences ($\mathrm{y_{i-2}, y_{i-1}}$)
    \item Previous-Next Source (\textbf{P-N-SRC}): Previous and next source sentences ($\mathrm{x_{i-1}, x_{i+1}}$)
\end{itemize}

\section{Experiment Setup}
\label{sec: experimental_setup}

We train our models with the proposed cascade MTL approach. The model is trained on $\langle\mathrm{c_x}, \mathrm{x}, \mathrm{y}\rangle$ triplet to jointly optimize both translation and source reconstruction objectives (Figure:~\ref{fig:model}). We also train three other contrastive models to show the effect of context in the MTL setting.
\paragraph{Vanilla-Sent:} A vanilla sentence-level baseline model is trained without context on a single encoder-decoder network.
\paragraph{Concat-Context:} This model is trained on a single encoder-decoder network where context is concatenated with the source \cite{tiedemann-scherrer-2017-neural,agrawal2018contextual,junczys-dowmunt-2019-microsoft} and fed to the encoder as input. In this setting, sentences within the context are concatenated with a unique token, `$\langle\mathrm{break}\rangle$'. The context and the source are concatenated with another special symbol, `$\langle\mathrm{concat}\rangle$'. The special symbol helps the model to distinguish between context and source sentences.
\paragraph{Inside-Context:} We re-implemented the `Inside-Context' model proposed by Li et al.~\shortcite{li-etal-2020-multi-encoder}, a multi-encoder approach. This model consists of two encoders and one decoder. The decoder is modified to include two cross-attention layers to attend over the outputs of both encoders before passing through the position-wise feed-forward layer \cite{vaswani2017attention}.

\subsection{Data Statistics}
We conduct experiments on WMT news-commentary, IWSLT`17 TED, and Europarl-v7 German-English corpora. For the WMT news-commentary, we use news-commentary v14 \cite{barrault-etal-2019-findings}\footnote{\url{https://data.statmt.org/news-commentary/v14/training/}} as the train set, newstest2017 as the validation set, and newstest2018 as the test set. For IWSLT`17 TED and Europarl-v7 corpora, we follow the train, validation, and test set splits mentioned in \cite{maruf-etal-2019-selective}\footnote{\url{https://github.com/sameenmaruf/selective-attn/tree/master/data}}. All models are trained on German to English. Table~\ref{tab: corpora_stats} shows data statistics of the train, validation, and test sets.

\begin{table}[!ht]
    \centering
    \resizebox{1.0\linewidth}{!}{
    \begin{tabular}{lcc}
        \toprule
        \textbf{Data} & \textbf{\# Sent} & \textbf{\# Doc} \\
        \cmidrule{1-3}
        News & 329,000/3,004/2,998 & 8,462/130/122 \\
        TED & 206,112/8,967/2,271 & 1,698/93/23 \\
        Europarl & 1,666,904/3,587/5,134 & 117,855/240/360 \\
        \bottomrule
    \end{tabular}
    }
    \caption{Data statistics for our experiments. \textbf{\# Sent}, \textbf{\# Doc} represent the number of sentences and documents, respectively. The numbers are shown in the Train/Validation/Test set order.}
    \label{tab: corpora_stats}
\end{table}

\begin{table*}[!ht]
    \centering
    \small
    \begin{tabular}{lcccccccc}
        \toprule
        \multirow{2}*{\textbf{Model}} & \multicolumn{2}{c}{\textbf{News}} & & \multicolumn{2}{c}{\textbf{TED}} & & \multicolumn{2}{c}{\textbf{Europarl}} \\
        \cmidrule{2-3}\cmidrule{5-6}\cmidrule{8-9}
         & s-BLEU & d-BLEU & & s-BLEU & d-BLEU & & s-BLEU & d-BLEU \\
        \cmidrule{1-9}
        Vanilla-Sent & 18.3 & 20.9 & & 19.9 & 24.9 & & 32.3 & 35.1 \\
        \cmidrule{1-9}
        Concat-Context: P@2-SRC & 18.0 & 20.5 & & 17.3 & 22.4 & & 32.5 & 35.4 \\
        Concat-Context: P-N-SRC & 18.4 & 20.7 & & 17.5 & 22.5 & & 32.7 & 35.6 \\
        Concat-Context: P@2-TGT & 14.7 & 17.2 & & 15.3 & 20.4 & & \textbf{36.4} & \textbf{39.1} \\
        \cmidrule{1-9}
        MTL: P@2-SRC & 19.1 & 21.7 & & 20.2 & 24.8 & & 29.5 & 32.6 \\
        MTL: P-N-SRC & $\textbf{20.1}^{\dag}$ & \textbf{22.5} & & 20.3 & 25.2 & & $32.5^{\dag}$ & 35.3 \\
        MTL: P@2-TGT & 19.2 & 21.7 & & $\textbf{20.7}^{\dag}$ & \textbf{25.4} & & 28.2 & 31.6 \\
        \bottomrule
    \end{tabular}
    \caption{BLEU scores of Vanilla-Sent, Concat-Context, and proposed MTL DocNMT models trained with different source contexts for German to English direction on News-commentary v14, IWSTL-17 TED, and Europarl corpora. \textbf{s-BLEU} and \textbf{d-BLEU} represent sentence-level and document-level BLEU respectively. The best results are shown in bold. `$\dag$' denotes the statistically significant results than Vanilla-Sent and Concat-Context models with $p < 0.05$.}
    \label{tab: main_results}
\end{table*}

\subsection{NMT Model Setups}
We conduct all the experiments on transformer architecture \cite{vaswani2017attention}. All the models are implemented in PyTorch\footnote{\url{https://pytorch.org/}}. We use 6-layer encoder-decoder stacks with 8 attention heads. Positional token embedding sizes are set to 512, and the feed-forward layer consists of 2048 cells. Adam optimizer \cite{DBLP:journals/corr/KingmaB14} is used for training with a noam learning rate scheduler \cite{vaswani2017attention} with an initial learning rate of 0.2. We use warmup steps of 16,000 \cite{popel2018training}, and dropout is set to 0.1. Due to the GPU memory restrictions, we use a mini-batch of 40 sentences for the models trained on News and TED corpora and 25 for the models trained on Europarl corpus. We create joint subword vocabularies of size 32k for each training corpus. We use the BPE \cite{sennrich-etal-2016-neural} to create subword vocabularies with SentencePiece \cite{kudo-richardson-2018-sentencepiece} implementation. We also learn the positional encoding of tokens \cite{devlin-etal-2019-bert}, and the maximum sequence length is set to 140 tokens for all models and 160 for \textit{Concat-Context} models.

All the models are trained till convergence. We use the perplexity of the validation set as an early stopping criterion with the patience of 10 \cite{popel2018training}. We report results on the best model checkpoint saved during the training. We perform beam search during inference with beam size 4 and length penalty of 0.6 \cite{45610}. For DocNMT models, we use the same source context with which the models are trained. Since the input to the intermediate decoder (source sentence) is also given during the testing phase, the representation of the intermediate decoder can be calculated in parallel, similar to the training phase.

All the experiments are conducted on a single Nvidia GTX 2080ti GPU. The number of parameters and training time of the models is as follows: \textit{Vanilla-Sent}: 76M, 76.5 hours, \textit{Concat-Context}: 76M, 81 hours, \textit{Inside-Context}: 118M, 125 hours and proposed \textit{MTL}: 130M, 160 hours. The parameters and training times are approximately the same for all the corpora.

\section{Results and Analysis}
\label{sec: results_and_analysis}
This section discusses the results of the trained models and the context's effect on Multi-Encoder and MTL settings. Table~\ref{tab: main_results} shows the sentence-BLEU (s-BLEU) and document-BLEU (d-BLEU) \cite{liu-etal-2020-multilingual-denoising,bao-etal-2021-g} scores of the proposed multi-task learning model along with the \textit{Vanilla-Sent} and \textit{Concat-Context} models.

We report all models' BLEU scores on German $\rightarrow$ English direction, calculated with sacreBLEU \cite{post-2018-call}. 

\subsection{Results of MTL and Contrastive Models}
We report the BLEU scores of the models on German $\rightarrow$ English direction, calculated with sacreBLEU \cite{post-2018-call}\footnote{sacreBLEU signature:``nrefs:1$|$case:mixed$|$eff:no$|$tok:13a$|$\\smooth:exp$|$version:2.3.1''}. The proposed MTL model can outperform both \textit{Vanilla-Sent} and \textit{Concat-Context} models by achieving s-BLEU scores of 20.1 (\textit{MTL: P-N-SRC}) and 20.7 (\textit{MTL: P@2-TGT}) with an improvement of +1.8 and +0.8 BLEU improvement for News and TED corpora respectively. However, in the case of the Europarl data set, \textit{Concat-Context} models outperform both \textit{Vanilla-Sent} and \textit{MTL} models. This shows that the \textit{Concat-Context} model requires more data to perform well, unlike the MTL models, which can also work effectively in low-resource settings. We observe that the performance of the models is almost uniform across the three different context settings with a maximum BLEU difference of +1.0 (\textit{P-N-SRC} vs. \textit{P@2-SRC}) on News, +0.5 (\textit{P@2-TGT} vs. \textit{P@2-SRC}) on TED and +4.3 (\textit{P-N-SRC} vs \textit{P@2-TGT}) on Europarl corpora respectively.

We also report d-BLEU (document-level BLEU) scores \cite{liu-etal-2020-multilingual-denoising,bao-etal-2021-g} by converting each document into one single sequence (paragraph) by concatenating all sentences from that document and calculate BLEU scores on the resulting corpus. This results in slightly higher scores than the sentence level by matching n-grams over the whole document instead of at the sentence level. Table~\ref{tab: main_results} also shows d-BLEU scores. Like s-BLEU scores, proposed MTL models achieve the best d-BLEU scores of 22.5 and 25.4 for News and TED corpora, respectively. We report the paired bootstrap resampling \cite{koehn-2004-statistical} results, calculated with sacreBLEU \cite{post-2018-call}.

\begin{table}[!ht]
    \centering
    \small
    \begin{tabular}{lccc}
        \toprule
        \textbf{Model} & \textbf{News} & \textbf{TED} & \textbf{Europarl} \\
        \cmidrule{1-4}
        MTL: P@2-SRC & 1.3 & 1.4 & 4.9 \\
        MTL: P@2-TGT & 1.2 & 1.6 & 3.9 \\
        MTL: P-N-SRC & 1.3 & 1.5 & 3.1\\
        \bottomrule
    \end{tabular}
    \caption{s-BLEU scores for the reconstruction objective of the MTL models on test set for News, TED, and Europarl corpora.}
    \label{tab: reconstruction_bleu}
\end{table}

\subsection{Analysis of Reconstruction Objective}
We analyze the performance of the MTL model on the reconstruction objective on the test set to verify if the context encoder is generating noise. If the context encoder generates noise by the suboptimal encoding of context, the intermediate decoder will fail to reconstruct the source sentence from the context; otherwise, the intermediate decoder can reconstruct the source sentence to a similar extent as the final translated sentence. We perform greedy decoding on the intermediate decoder to generate the source from the context. Table~\ref{tab: reconstruction_bleu} shows the BLEU scores of the reconstruction objective on the test set for News, TED, and Europarl corpora. The results show that the MTL models fail to reconstruct the source from the context. Based on this, we conclude that the context encoder cannot encode the context, leading to poor reconstruction performance of the models. However, we hypothesize that the model cannot reconstruct the source from the context because the corpora used to train context-aware models might not be context-aware. This observation aligns with the previous works \cite{kim-etal-2019-document,li-etal-2020-multi-encoder}, and with enough data, vanilla sentence-level NMT models can outperform the document-level NMT models.

\begin{figure}[!ht]
    \centering
    \includegraphics[scale=0.15]{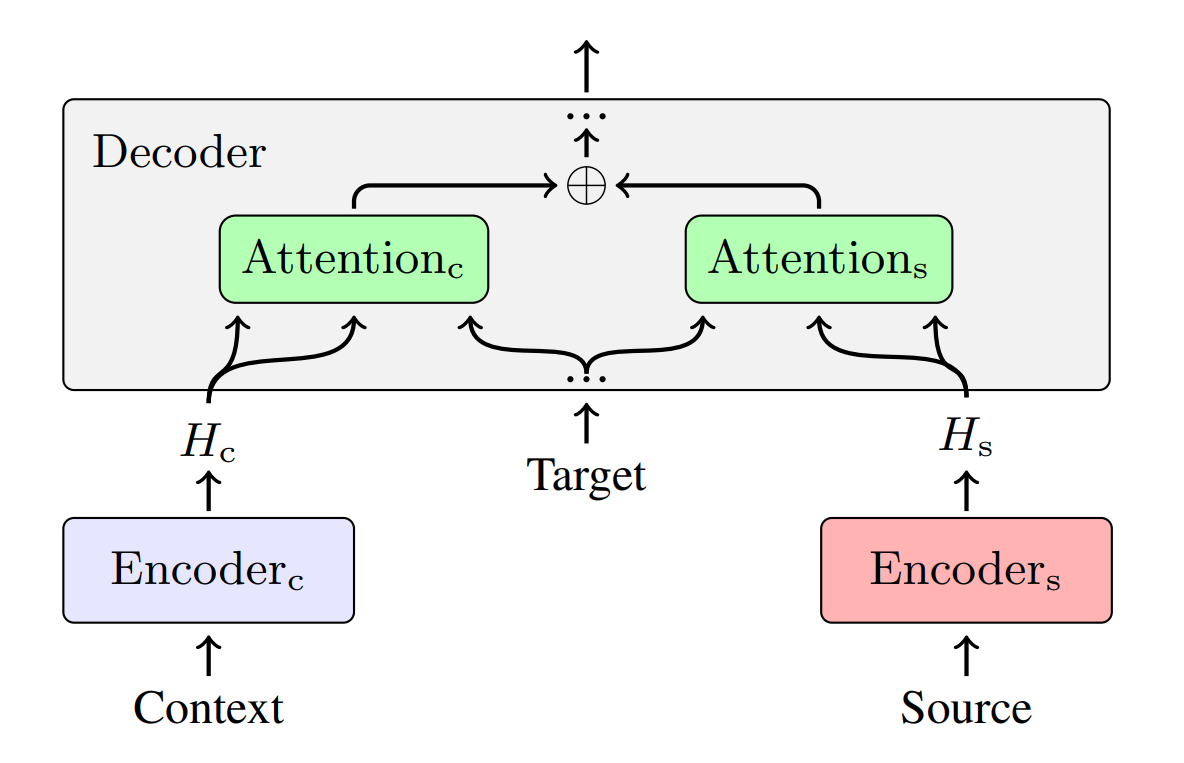}
    \caption{The overview of the Inside-Context model. The input to the model is a triplet consisting of (\textit{Context, Source, Target}). The multi-head attention layer of the decoder is modified to attend to both the context encoders ($\mathrm{Encoder_c}$) and the source encoder ($\mathrm{Encoder_s}$).}
    \label{fig: inside_attn}
\end{figure}

\begin{table}[!ht]
    \centering
    \resizebox{1.0\linewidth}{!}{
    \begin{tabular}{lccc}
        \toprule
        \textbf{Model} & \textbf{News} & \textbf{TED} & \textbf{Europarl} \\
        \cmidrule{1-4}
        MTL: P@2-SRC & 19.1 & 20.2 & 29.5 \\
        MTL: P-N-SRC & $\textbf{20.1}^\dag$ & 20.3 & 32.5$^\dag$ \\
        MTL: P@2-TGT & 19.2 & $\textbf{20.7}^\dag$ & 28.2 \\
        \cmidrule{1-4}
        Inside-Context: P@2-SRC & 18.8 & 19.6 & 33.2 \\
        Inside-Context: P-N-SRC & 19.0 & 19.8 & 33.2 \\
        Inside-Context: P@2-TGT & 18.3 & 20.4 & \textbf{33.6} \\
        \bottomrule
    \end{tabular}
    }
    \caption{Comparison of s-BLEU scores of MTL and Inside-Context Multi-Encoder models. The best results are shown in bold. `$\dag$' denotes the statistically significant results than Vanilla-Sent and Concat-Context models with $p < 0.05$.}
    \label{tab: multi-encoder_multi-decoder_bleu}
\end{table}

\subsection{MTL vs. Multi-Encoder Approach}
We compare the proposed MTL approach to the existing Multi-Encoder approach to study how the model will perform in a single-task setting. Specifically, we compare our MTL approach (single-encoder multi-decoder network) with \textit{Inside-Context} \cite{li-etal-2020-multi-encoder} architecture. This model consists of two transformer encoders and one transformer decoder. Figure~\ref{fig: inside_attn} shows the model's architecture. The decoder is modified to attend to the outputs of both encoders. The model follows the transformer \cite{vaswani2017attention} architecture. An element-wise addition is performed on the outputs of both cross-attention layers before passing through layer-norm and position-wise feed-forward layers. Table~\ref{tab: multi-encoder_multi-decoder_bleu} shows the s-BLEU scores of the MTL and Inside-Context models. We observe that the performance of multi-encoder models is similar to MTL models, with MTL models achieving +1.1 (P-N-SRC models), +0.3 (P@2-TGT models) BLEU points improvement over Inside-Context models for News and TED corpora respectively. In the case of Europarl, inside-context models achieve better performance than the MTL models, with the P@2-TGT model achieving +5.4 BLEU points improvement compared to the MTL model. Based on the results, we conclude that the MTL setting is more effective for low-resource scenarios.

\begin{table}[!ht]
    \centering
    \resizebox{1.0\linewidth}{!}{
    \begin{tabular}{lccc}
        \toprule
        \textbf{Model} & \textbf{News} & \textbf{TED} & \textbf{Europarl} \\
        \cmidrule{1-4}
        MTL: P@2-SRC & 1.2 (-17.9) & 0.8 (-19.4) & 4.5 (-25.0) \\
        MTL: P-N-SRC & 1.2 (-18.9) & 0.8 (-19.5) & 4.0 (-28.5) \\
        MTL: P@2-TGT & 0.5 (-18.7) & 0.3 (-20.4) & 3.9 (-24.3) \\
        \cmidrule{1-4}
        Inside-Context: P@2-SRC & 18.7 (-0.1) & 19.4 (-0.2) & 33.2 (0.0) \\
        Inside-Context: P-N-SRC & 18.9 (-0.1) & 19.8 (0.0) & 33.2 (0.0) \\
        Inside-Context: P@2-TGT & 18.3 (0.0) & 20.3 (-0.1) & 33.1 (-0.5) \\
        \bottomrule
    \end{tabular}
    }
    \caption{Comparison of s-BLEU scores of MTL models tested with random context. The difference in scores over the models trained with the selected context is shown inside the parentheses.}
    \label{tab: random_context_scores}
\end{table}

\subsection{Effect of Context in MTL setting}
\label{sub_sec: effect_of_context_in_mtl}
Since the BLEU scores of our MTL models are almost the same for all three context settings, we check whether the MTL models are affected by the choice of context. To this end, we test the MTL models with random context. Here, random context denotes two randomly selected sentences from the entire corpus. Table~\ref{tab: random_context_scores} shows the results of MTL and Inside-Context models tested with random context. Results show that the MTL models fail to translate source sentences when the context is random. However, Inside-Context models are agnostic to context as models can translate well even if the context is random. Our findings in the case of multi-encoder models are in line with the findings of Li et al.~\cite{li-etal-2020-multi-encoder}. Based on the results, we conclude that MTL models are sensitive to the choice of context. Section~\ref{sec: effect_of_random_context} describes a similar experiment where the MTL models are tested with random context. However, the architecture used in the main experiments differs slightly from the one used in the preliminary investigation. We observe that feeding the Intermediate Decoder output to the Final Decoder makes the model sensitive to the choice of context (cf. Figure~\ref{fig:model} and Figure~\ref{fig:initial_model} in the Appendix~\ref{sec: aux_objectives}). We hypothesize that a weighted combination of the Context Encoder output and Intermediate Decoder output is desired as it performs slightly better than the model used in the main experimental setup. However, it also makes the model agnostic to the choice of context. We plan to explore this behaviour in detail in our future work.

\begin{table}[!ht]
    \centering
    \resizebox{1.0\linewidth}{!}{
    \begin{tabular}{lccc}
        \toprule
        \textbf{Model} & \textbf{News} & \textbf{TED} & \textbf{Europarl} \\
        \cmidrule{1-4}
        MTL: P@2-SRC & 13.7 (+12.5) & 11.2 (+10.4) & 22.3 (+17.8) \\
        MTL: P-N-SRC & 14.5 (+13.3) & 11.3 (+10.5) & 19.7 (+15.7) \\
        \cmidrule{1-4}
        Inside-Context: P@2-SRC & 18.7 (0.0) & 19.6 (+0.2) & 33.1 (-0.1) \\
        Inside-Context: P-N-SRC & 19.0 (+0.1) & 19.7 (-0.1) & 33.0 (-0.2) \\
        \bottomrule
    \end{tabular}
    }
    \caption{s-BLEU scores of the MTL and Inside-Context models are tested by giving the same source sentences as context and input. The change of s-BLEU scores over the models tested with random context is shown in ($\pm x$).}
    \label{tab: sst_bleu}
\end{table}

\subsection{Results of MTL and Multi-Encoder models without Context}
We conduct experiments on MTL and Inside-Context models by using the same source sentence as the context. Since the proposed MTL models fail when tested with random context (cf. Section~\ref{sub_sec: effect_of_context_in_mtl}), we observe how the MTL and Multi-Encoder models are performing when the same source sentence is given as context. This setting presents a scenario where the context is not random but also not the type of context with which the models are trained. We conduct experiments for \textit{P@2-SRC} and \textit{P-N-SRC} context settings only as the \textit{P@2-TGT} context setting requires the current target sentence, which is unavailable during testing. We observe that MTL models can perform well compared to the random context setting, which shows that the MTL models are sensitive to the choice of context. The performance of Inside-Context models is almost the same as those tested with random context. This shows that the Inside-Context model is agnostic to the choice of the context. Table~\ref{tab: sst_bleu} shows the s-BLEU scores of the MTL and Inside-Context models.

\begin{table}[!ht]
    \centering
    \resizebox{1.0\linewidth}{!}{
    \begin{tabular}{lccc}
        \toprule
        \textbf{Model} & \textbf{News} & \textbf{TED} & \textbf{Europarl} \\
        \cmidrule{1-4}
        Vanilla-Sent & 40.17 & 31.22 & 37.22 \\
        \cmidrule{1-4}
        Concat-Context: P@2-SRC & 39.34 & 30.01 & 36.42 \\
        Concat-Context: P-N-SRC & 39.99 & 29.57 & 36.78 \\
        Concat-Context: P@2-TGT & 38.50 & 28.82 & \textbf{37.27} \\
        \cmidrule{1-4}
        MTL: P@2-SRC & 40.69 & 31.44 & 35.96 \\
        MTL: P-N-SRC & 40.50 & 31.24 & 36.94 \\
        MTL: P@2-TGT & \textbf{40.99} & \textbf{31.90} & 33.91 \\
        \bottomrule
    \end{tabular}
    }
    \caption{Accuracy of Pronoun Translation (APT) scores. The best results are shown in bold.} 
    \label{tab: apt_scores}
\end{table}

\subsection{Pronoun Translation Accuracy}
We also evaluate our proposed models' performance on pronoun translation accuracy. We calculate the pronoun translation accuracy with APT (accuracy of pronoun translation) \cite{miculicich-werlen-popescu-belis-2017-validation} metric\footnote{\url{https://github.com/idiap/APT}}. This metric requires a list of pronouns from the source language (German) with a list of pronouns from the target language (English) as an optional argument. We use spaCy\footnote{\url{https://spacy.io/models}} to tag both source and target sentences from the test set and extract pronouns. Table \ref{tab: apt_scores} shows the APT scores of \textit{Vanilla-Sent}, \textit{Concat-Context}, and \textit{MTL} DocNMT models. The APT scores correlate with the s-BLEU and d-BLEU scores, achieving the highest APT score of 40.99 in \textit{MTL: P@2-TGT} setting with an improvement of +0.82 over \textit{Vanilla-Sent} and +1.0 over \textit{Concat-Context} (\textit{P-N-SRC}) models on News corpus. Similarly, the \textit{MTL: P@2-TGT} model achieves the highest APT score of 31.90 with an improvement of +0.68 and +1.89 over \textit{Vanilla-Sent} and \textit{Concat-Context} (\textit{P@2-SRC}) on TED. For the Europarl corpus, \textit{Concat-Context} (\textit{P@2-TGT}) achieved the highest APT score of 37.27 with an improvement of +0.05 and +0.33 over \textit{Vanilla-Sent} and \textit{MTL} (\textit{P-N-SRC}) models respectively.

\section{Conclusion}
\label{sec: conclusion}
This work explored the MTL approach for document-level NMT (DocNMT). Our proposed MTL approach is based on cascade MTL architecture, where the model consists of one encoder (for context encoding) and two decoders (for the representation of the current source and target sentences). Reconstruction of the source sentence given the context is considered the auxiliary task, along with the translation of the current source sentence as the main task. We conducted experiments for German--English for News-commentary v14, IWSLT`17 TED, and Europarl v7 corpora with three different types of contexts \textit{viz.} two previous sources, two previous targets, and previous-next source sentences with respect to the current input source sentence.

Our proposed MTL approaches outperform the sentence-level baseline and concatenated-context models in low-resource (for News and TED corpora) settings. However, all models perform well in the high resource setting (Europarl corpus), with proposed MTL models slightly underperforming the rest. Our MTL models are more sensitive to the choice of context than the multi-encoder models when tested with random context. We observe that the context encoder cannot encode context sufficiently and performs poorly reconstruction tasks. Finally, we reported APT (accuracy of pronoun translation) scores, and the proposed MTL models outperformed the sentence-level baseline and concatenated-context models. Our empirical analysis concludes that our approach is more sensitive to the choice of context and improves the overall translation performance in low-resource context-aware settings. We plan to explore other tasks, such as gap sentence generation (GSG) \cite{zhang2020pegasus} as an auxiliary task for better context encoding, different training curricula to prioritize one objective over the other during the training, and dynamic context selection.

\section{Limitations}
Our study poses two main limitations. First, our primary motivation is to understand the effect of context and if the context encoding can be modelled as an auxiliary task but not to propose a model to achieve state-of-the-art results. We have followed the findings of Li et al.~\cite{li-etal-2020-multi-encoder} and used one of their approach to understanding the effect of context. Our observations are also in line with their findings.

Second, even though our proposed MTL approach can outperform the models in other settings, the auxiliary task (reconstruction) is not very effective as it improves the BLEU scores in the range of [0.1-1.8] over the Multi-encoder models. We hypothesize that, in the loss function, we are giving equal weights to both the objectives (0.5 for both reconstruction and translation objectives), which might lead to significantly less improvement in overall translation quality. We plan to explore different training curricula to adjust the weight of the objectives dynamically during the training.

\section*{Acknowledgements}
We gratefully acknowledge the support from the “NLTM: VIDYAAPATI” project, sponsored by Electronics and IT, Ministry of Electronics and Information Technology (MeiTY), Government of India. Santanu Pal acknowledges the support from Wipro AI. We also thank the anonymous reviewers for their insightful comments.


\bibliography{eamt24,anthology_new,custom_new}
\bibliographystyle{eamt24}

\appendix
\section{Appendix}
\subsection{Preliminary Investigation on Auxiliary Objectives}
\label{sec: aux_objectives}

The joint probability in Equation~\ref{eq:mtl_obj} ($p(\mathrm{x}, \mathrm{c_x}$)) can be calculated in two ways such as:

\begin{eqnarray}
    p(\mathrm{x}, \mathrm{c_x}) = p(\mathrm{x}|\mathrm{c_x}) \times p(\mathrm{c_x}) \\
    p(\mathrm{x}, \mathrm{c_x}) = p(\mathrm{c_x}|\mathrm{x}) \times p(\mathrm{x})
\end{eqnarray}

Since the joint probability can be computed in two different ways, we conduct an initial study to select the optimal auxiliary objective that improves the overall translation performance of the model. Specifically, we consider $p(\mathrm{x}|\mathrm{c_x})$ as one auxiliary task where source ($\mathrm{x}$) is autoregressively reconstructed (denoted as \textbf{Re-Src}) from the encoded context ($\mathrm{c_x}$) and $p(\mathrm{c_x}|\mathrm{x})$ as the other auxiliary task where context ($\mathrm{x}$) is autoregressively reconstructed (denoted as \textbf{Re-Cntx}) from the encoded source ($\mathrm{c_x}$. We conducted experiments to verify which auxiliary task is performing better.

\begin{figure}[!ht]
    \centering
    \includegraphics[scale=0.5]{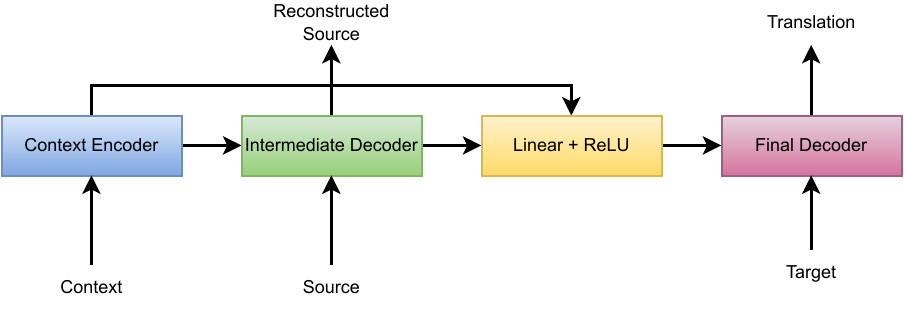}
    \caption{The overview of modified MTL architecture with residual connection. The input to the model is a triplet. The triplet consist of (\textit{Context}, \textit{Source}, \textit{Target}) in \textbf{Re-Src} setting and (\textit{Source}, \textit{Context}, \textit{Target}) in \textbf{Re-Cntx} setting. Here, \textit{Source}: Current source sentence, \textit{Context}: Context for the current source sentence, and \textit{Target}: Translation of current source sentence. None of the layers are shared.}
    \label{fig:initial_model}
\end{figure}

The experimental setup and model architecture are slightly different for this comparison study than those used in the main experiments.\footnote{We modified the experimental setup and model architecture during our main experiments. In this preliminary investigation, the capacity of models with independent subword vocabularies is slightly larger. Due to this, the s-BLEU scores are slightly better than the main results.} The Context Encoder and Intermediate Decoder output are combined with a linear layer with ReLU activation. The main experimental setup does not use this linear layer + ReLU combination. We hypothesize that adding this layer might make the model agnostic to the choice of context. We test this by training the model with random context (cf. Section~\ref{sec: effect_of_random_context}. Specifically, we use two context settings \textit{viz.} \textit{P@2-SRC} and \textit{P-N-SRC} settings (cf. \ref{subsec: src_cntx}). We use a fixed learning rate of $10^{-5}$ instead of the warmup schedule. The output from this layer is given as input to the Final Decoder.

\begin{table}[!ht]
    \centering
    \resizebox{1.0\linewidth}{!}{
    \begin{tabular}{ccccc}
        \toprule
        \multicolumn{2}{c}{\textbf{Model}} & \textbf{Vanilla-Sent} & \textbf{MTL: P@2-SRC} & \textbf{MTL: P-N-SRC} \\
        \cmidrule{1-5}
        \multirow{2}*{News} & Re-Src & \multirow{2}*{16.5} & 20.6 & \textbf{20.9} \\
                            & Re-Cntx &  & 16.7 (-3.9) & 17.9 (-3.0) \\
        \cmidrule{1-5}
        \multirow{2}*{TED} & Re-Src & \multirow{2}*{12.1} & 21.6 & \textbf{22.0} \\
                           & Re-Cntx & & 18.0 (-3.6) & 17.8 (-4.2) \\
        \cmidrule{1-5}
        \multirow{2}*{Europarl} & Re-Src & \multirow{2}*{35.0} & 35.1 & \textbf{35.8} \\
                                & Re-Cntx &  & 33.2 (-1.9) & 33.6 (-2.2) \\
        \cmidrule{1-5}
    \end{tabular}
    }
    \caption{Comparison of s-BLEU scores of Baseline and proposed MTL DocNMT models trained with different source contexts for German to English direction. Differences in the scores over \textbf{Re-Src} are shown inside the parentheses.}
    \label{tab: appendix_results}
\end{table}

We use a mini-batch of 18 sentences to train all the models. We create two separate subword vocabularies for each training corpus. The created subword vocabulary is 40k in both German and English. We use the unigram language model \cite{kudo-2018-subword} to create subword vocabularies with SentencePiece \cite{kudo-richardson-2018-sentencepiece}, and the maximum sequence length is set to 160 tokens. During inference, we perform greedy decoding. The rest of the experimental setup is the same as the one used in the main experiments. 

\begin{table}[!ht]
    \centering
    \resizebox{1.0\linewidth}{!}{
    \begin{tabular}{lccccc}
        \toprule
        \multirow{2}*{\textbf{Model}} & \multicolumn{2}{c}{\textbf{Random-Train}} & & \multicolumn{2}{c}{\textbf{Random-Infer}} \\
        \cmidrule{2-3}\cmidrule{5-6}
         & Re-Src & Re-Cntx & & Re-Src & Re-Cntx \\
        \cmidrule{1-6}
        MTL: P@2-SRC & 20.9 & 16.6 & & 20.6 & 16.8 \\
        MTL: P-N-SRC & 20.9 & 16.4 & & 20.8 & 17.8 \\
        \bottomrule
    \end{tabular}
    }
    \caption{s-BLEU scores of \textit{Random-Train} and \textit{Random-Infer} experiments on News-commentary corpus.}
    \label{tab: effect_of_random_context}
\end{table}

\subsubsection{Effect of Random Context}
\label{sec: effect_of_random_context}
We also conduct experiments to study how the random context affects the MTL models. Specifically, we evaluate the MTL models in two settings. The model is trained on the random context in the \textit{Random-Train} setting by concatenating two randomly sampled sentences from the train set and testing with\textit{ P@2-SRC} and \textit{P-N-SRC} context settings. In \textit{Random-Infer} setting, the model is trained on \textit{P@2-SRC} and \textit{P-N-SRC} context settings and tested with random context. We train the models on the news-commentary corpus. Table~\ref{tab: effect_of_random_context} shows the s-BLEU scores of the MTL models trained and tested in the random context setting. Based on the results, we conclude that the model trained with random context improves the robustness of the model. This observation aligns with the findings of Li et al.~\shortcite{li-etal-2020-multi-encoder}, but they conducted experiments in the non-MTL setting with multiple encoders. As the model largely ignores the choice of the context, we remove this linear + ReLU combination and feed the output of the Intermediate Decoder to the Final Decoder. We hypothesize that this forces the model to consider the context while generating the target sentence.

\end{document}